\documentclass[10pt]{article}
\usepackage{rev_sbrn}
\usepackage{times,amsfonts,enumerate,amssymb,amsmath,epsfig,subfigure,bm,cite,multirow}
\usepackage[latin1]{inputenc}
\usepackage[OT1]{fontenc}
\usepackage[brazil]{babel}

\usepackage[a4paper, 
hmargin={2cm,1cm}, 
vmargin={2cm,2cm},
footskip=5mm]{geometry}

\begin{document}
\title{TRIAGEM VIRTUAL DE IMAGENS DE IMUNO-HISTOQUÍMICA USANDO REDES NEURAIS ARTIFICIAIS E ESPECTRO DE PADRÕES}

\author{%
{\bf Higor Neto Lima} \\ 
{\normalsize Escola Politécnica de Pernambuco, Universidade de Pernambuco}\\ 
{\normalsize higornetto@gmail.com} \\
{\bf Wellington Pinheiro dos Santos} \\ 
{\normalsize Núcleo de Engenharia Biomédica, Centro de Tecnologia e Geociências, Universidade Federal de Pernambuco}\\ 
{\normalsize wellington.santos@ufpe.br} \\
{\bf Mêuser Jorge Silva Valença}\\
{\normalsize Escola Politécnica de Pernambuco, Universidade de Pernambuco}\\ 
{\normalsize mjv@ecomp.poli.br} \\
}
\maketitle 

\thispagestyle{cbrna}
\pagestyle{cbrna}

\setcounter{page}{1}

%
\noindent{{\bf\large Resumo --} A importância de se organizar imagens médicas de acordo com sua natureza, aplicação e relevância tem aumentado. Além do mais, a seleção prévia de imagens médicas pode ser bastante útil para reduzir o tempo de análise dispendido por patologistas. Neste artigo é proposto um classificador de imagens para integrar um sistema de CBIR (\emph{Content-Based Image Retrieval}, Recuperação de Imagens Baseada em Conteúdo) dedicado à seleção prévia de imagens, ou seja, à tarefa de triagem. Esse classificador é baseado na representação das imagens de entrada pelo espectro de padrões e na classificação por redes neurais. A seleção dos atributos é realizada aplicando a Análise de Componentes Principais aos vetores de atributos obtidos pela extração do espectro de padrões, enquanto a classificação de imagens é baseada no uso de redes neurais perceptron multicamadas e na composição de mapas auto-organizados e aprendizado por quantização vetorial. Esses métodos foram aplicados à seleção baseada em conteúdo de imagens de imuno-histoquímica de placenta e pulmão de neomortos. Os resultados demonstraram que esta abordagem pode atingir desempenhos de classificação razoáveis.}
\\

\noindent{{\bf\large Palavras-chave --} CBIR, k-Médias, Algoritmos de Agrupamento, Análise de Componentes Principais, Espectro de Padrões, Imuno-Histoquímica.}
\\

\noindent{{\bf\large Abstract --} The importance of organizing medical images according to their nature, application and relevance is increasing. Furhermore, a previous selection of medical images can be useful to accelerate the task of analysis by pathologists. Herein this work we propose an image classifier to integrate a CBIR (Content-Based Image Retrieval) selection system. This classifier is based on pattern spectra and neural networks. Feature selection is performed using pattern spectra and principal component analysis, whilst image classification is based on multilayer perceptrons and a composition of self-organizing maps and learning vector quantization. These methods were applied for content selection of immunohistochemical images of placenta and newdeads lungs. Results demonstrated that this approach can reach reasonable classification performance.}
\\

\noindent{{\bf\large Keywords --} CBIR, k-means, clustering algorithms, principal componente analysis, pattern spectrum, immunohistochemistry.}
\\

\section{Introdução}

A velocidade de processamento dos computadores e a capacidade de armazenamento, bem como o desenvolvimento de novas mídias de armazenamento e evolução das mídias existentes, têm passado por um crescimento vertiginoso desde o começo da década de 1990 \cite{gagaudakis2003,gagaudakis2002}. Paralelamente, a evolução das tecnologias digitais tem sido cada vez mais empregada no desenvolvimento de soluções baseadas em processamento, armazenamento e análise de imagens digitais, o que, por sua vez, tem demandado mais capacidade de armazenamento e representação de imagens \cite{castanon2003}. Aliada a essa demanda surge também a necessidade de construir métodos de busca que permitam a pesquisa de imagens baseada em conteúdo \cite{castanon2003}.

Recuperação de Imagens Baseada em Conteúdo (\emph{Content-Based Image Retrieval}, CBIR) é uma forma de buscar imagens por meio de uma imagem-base, selecionada de acordo com o interesse do usuário. Para proporcionar a busca baseada em conteúdo, é preciso, portanto, representar as imagens no banco de dados de forma alternativa. Isso é feito usualmente por meio da construção de vetores de características que contenham os atributos de interesse para descrição e representação unívoca do conteúdo de interesse. As características mais usadas são cor, textura e forma. A recuperação de imagens baseada em conteúdo se fundamenta em três premissas básicas: extração de características visuais, indexação multidimensional, e projeto de sistemas de recuperação \cite{rui1999}.

A representação das imagens nos bancos de dados por meio de vetores de características proporciona que o usuário possa buscar imagens na base de dados com o contexto requerido fazendo uso de uma imagem base, representada por seu vetor de características. Esse vetor é comparado aos vetores que representam as outras imagens no banco de dados e, por meio de medidas de similaridade, a pesquisa retorna as imagens de conteúdo similar \cite{castanon2003,rui1999,zhang2003,huang2003,cai2003,long2003,jin2003}. Neste trabalho as imagens são representadas por vetores de características obtidos da extração do espectro morfológico de padrões de cada uma das bandas da imagem colorida \cite{ulisses1994}. A dimensionalidade desses vetores é então reduzida pela aplicação da Análise de Componentes Principais (\emph{Principal Component Analysis}, PCA).

O sistema de triagem virtual proposto neste trabalho é apresentado em duas versões: uma primeira, baseada em redes neurais MLP (\emph{Multilayer Perceptron}), e uma segunda versão, baseada em redes SOM (\emph{Self-Organizing Maps}) e redes LVQ (\emph{Learning Vector Quantization}). As duas configurações são comparadas quanto ao desempenho na tarefa de triagem virtual, ou seja, quanto à taxa de acerto de pré-classificação. A escolha pelo uso de redes neurais se deve ao seu emprego em outros trabalhos envolvendo CBIR \cite{antani2003,seo2007}.

Este trabalho está organizado da forma que segue: na seção `Materiais e Métodos' apresentam-se a teoria básica das redes neurais artificiais do tipo perceptron multicamadas e mapas auto-organizados de Kohonen, os princípios de Morfologia Matemática e a definição do espectro de padrões, e a proposta de sistema de triagem virtual; em `Resultados Experimentais' são apresentados os resultados obtidos para as duas configurações de sistema de CBIR para triagem virtual. Por fim, na seção `Discussão e Conclusões' são apresentados comentários e conclusões gerais e específicas levando em conta os resultados obtidos.

\section{Materiais e Métodos} \label{sec:quanti_metodos}

\subsection{Redes Perceptron de Camada Única}

As redes perceptron consistem em classificadores lineares propostos por Rosenblatt que emprega neurônios MCP e uma técnica de treinamento supervisionado baseada no aprendizado de Hebb \cite{kovacs1996,braga2000}.

Em alguns textos tal rede é considerada uma rede de duas camadas de nodos \cite{kovacs1996}. Neste trabalho, quando se referir a uma rede de $M$ camadas, por exemplo, ficará explícito se são $M$ camadas de nodos ou de neurônios propriamente ditos, o que exclui a camada de nodos de entrada, que não executam funções matemáticas, mas apenas distribuem as entradas para a primeira camada de neurônios.

O treinamento de uma rede perceptron de camada única é baseado no seguinte procedimento, para uma rede de $m$ neurônios e $n$ atributos de entrada \cite{braga2000}:
\begin{enumerate}
    \item Inicializar os pesos $w_{i,j}(0)$.
    \item Inicializar a taxa de aprendizado $\eta$.
    \item Repetir até $\delta(t)\leq\epsilon$ ou um máximo de $N_{\max}$ iterações, onde $\epsilon$ é o máximo erro permitido:
    \begin{enumerate}
        \item Para cada par $(\mathbf{x},\mathbf{d})$ do conjunto de treinamento $\Psi$, onde
        $$
        \Psi = \{(\mathbf{x}^{(l)},\mathbf{d}^{(l)})\}_{l=1}^L,
        $$
        o vetor
        $$
        \mathbf{x}=(x_1, x_2, \dots, x_n)^T
        $$
        é o vetor de atributos de entrada e
        $$
        \mathbf{d}=(d_1, d_2, \dots, d_m)^T
        $$
        é o vetor de saídas desejadas dos $m$ neurônios, repetir:
        \begin{enumerate}
            \item Calcular as saídas $y_i(t)$ dos neurônios:
            $$ y_i(t)=g\left( \sum_{j=1}^n w_{i,j}(t)x_j(t) \right), $$
            onde $1\leq i\leq m$.
            \item Calcular o ajuste $\Delta w_{i,j}(t)$:
            $$ \delta_i(t)=d_i(t)-y_i(t), $$
            $$ \Delta w_{i,j}(t)=\eta \delta_i(t) x_j(t),   $$
            onde $1\leq i\leq m$ e $1\leq j\leq n$.
            \item Atualizar os pesos:
            $$ w_{i,j}(t+1)=w_{i,j}(t)+\Delta w_{i,j}(t), $$
            onde $1\leq i\leq m$ e $1\leq j\leq n$.
        \end{enumerate}
        \item Calcular $\delta(t)$ da forma que segue:
        $$ \delta(t)=\max(\delta_1(t), \delta_2(t),\dots, \delta_m(t)). $$
    \end{enumerate}
\end{enumerate}

As redes perceptron têm o inconveniente de apenas convergirem para casos linearmente separáveis, ou seja, casos em que existem hiperplanos que separam os conjuntos de entrada \cite{sklansky1981}. Essa limitação é resolvida com as redes perceptron multicamadas.

\subsection{Redes Perceptron Multicamadas}

As \emph{redes perceptron multicamadas} foram um avanço muito importante para o fim da estagnação das pesquisas em RNAs, devido à possibilidade de, utilizando neurônios MCP e arquitetura multicamadas, resolver problemas não-linearmente separáveis \cite{kovacs1996}. Elas utilizam um algoritmo chamado \emph{algoritmo de retropropagação}, que consiste em uma generalização do algoritmo de treinamento do perceptron de camada única através da estimação dos erros $\delta_i^{(k)}$ de uma camada $k$ a partir dos erros $\delta_i^{(k+1)}$ da camada superior $k+1$.

É possível provar que, para modelagem de funções, é possível modelar qualquer função contínua com uma rede perceptron de duas camadas de neurônios, enquanto qualquer função pode ser modelada usando um perceptron de três camadas de neurônios, diferente das redes de camada única, que só conseguem modelar funções contínuas e lineares \cite{braga2000}.

O algoritmo de retropropagação, também chamado de \emph{regra delta generalizada}, consiste no seguinte procedimento, para uma rede de $M+1$ camadas de nodos, cada uma com $N_k$ nodos, sendo $N_0=n$ atributos de entrada e $N_M=m$ neurônios de saída, onde
$0\leq k\leq M$ \cite{braga2000}:
\begin{enumerate}
    \item Inicializar os pesos $w_{i,j}^{(k)}(0)$.
    \item Inicializar a taxa de aprendizado $\eta$.
    \item Repetir até $\delta(t)\leq\epsilon$ ou um máximo de $N_{\max}$ iterações, onde $\epsilon$ é o máximo erro permitido:
    \begin{enumerate}
        \item Para cada par $(\mathbf{x},\mathbf{d})$ do con\-jun\-to de tre\-i\-na\-men\-to $\Psi$, onde
        $$
        \Psi = \{(\mathbf{x}^{(l)},\mathbf{d}^{(l)})\}_{l=1}^L,
        $$
        o vetor
        $$
        \mathbf{x}=(x_1, x_2, \dots, x_n)^T
        $$
        é o vetor de a\-tri\-bu\-tos de entrada e
        $$
        \mathbf{d}=(d_1, d_2, \dots, d_m)^T
        $$
        é o vetor de saídas desejadas dos $N(M)=m$ neurônios de saída, repetir:
        \begin{enumerate}
            \item Calcular as saídas $y_i^{(k)}(t)$ dos neurônios:
            $$ y_i^{(k)}(t)=g\left( \sum_{j=1}^{N_{k-1}} w_{i,j}^{(k)}(t)x_j^{(k)}(t) \right), $$
            onde $1\leq i\leq N_k$ e
            $$ x_j^{(k)}(t)=y_j^{(k-1)}(t), $$
            onde $1\leq k\leq M$ e
            $$ y_j^{(0)}(t)=x_j(t), $$
            para $1\leq j\leq N_{k-1}$, $N_0=n$ e $N_M=m$.
            \item Calcular os erros da camada de saída:
            $$ \delta_i^{(M)}(t)=(d_i(t)-y_i^{(M)}(t))g'(y_i^{(M)}(t)), $$
            onde $1\leq i\leq N_M$ e $g'(\centerdot)$ é a derivada de $g(\centerdot)$.
            \item Calcular os erros das outras camadas:
            $$ \delta_i^{(k)}(t)=g'(y_i^{(k)}(t)) \sum_{j=1}^{N_{k+1}} w_{j,i}^{(k+1)}(t) \delta_j^{(k+1)}(t), $$
            onde $1\leq i\leq N_k$ e $1\leq k\leq M-1$.
            \item Calcular os ajustes $\Delta w_{i,j}^{(k)}(t)$:
            $$ \Delta w_{i,j}^{(k)}(t)=\eta \delta_i^{(k)}(t) x_j^{(k)}(t), $$
            onde $1\leq i\leq N_k$, $1\leq j\leq N_{k-1}$ e $1\leq k\leq M$.
            \item Atualizar os pesos:
            $$ w_{i,j}^{(k)}(t+1)=w_{i,j}^{(k)}(t)+\Delta w_{i,j}^{(k)}(t), $$
            onde $1\leq i\leq N_k$, $1\leq j\leq N_{k-1}$ e $1\leq k\leq M$.
        \end{enumerate}
        \item Calcular $\delta(t)$ da forma que segue:
        $$ \delta(t)=\max(\delta_1^{(M)}(t), \delta_2^{(M)}(t),\dots, \delta_{N_M}^{(M)}(t)). $$
    \end{enumerate}
\end{enumerate}

As funções de ativação dos neurônios nas redes perceptron multicamadas devem ser contínuas, monotonicamente crescentes e
diferenciáveis \cite{arik2004}.

Existem outras versões do algoritmo de retropropagação onde os ajustes dos pesos são calculados segundo outros métodos, mas o princípio básico é o mesmo, estando a diferença apenas em questões como a rapidez da convergência e outras questões cuja natureza e interesse dependem da aplicação.

\subsection{Redes SOM}

Os \emph{mapas auto-organizados de Kohonen} foram originalmente desenvolvidos para fazer agrupamento de dados. Também chamados de redes SOM (\emph{Self-Organizing Maps}) \cite{braga2000}, os mapas auto-organizados possuem uma arquitetura bastante diferente das redes neurais usuais: neles os neurônios podem ser dispostos matricialmente, tentando emular a distribuição dos neurônios biológicos no cérebro.

Nas redes SOM os neurônios podem ser dispostos de forma linear ou matricial, em todas as dimensões possíveis, sendo que a quantidade de vizinhos influencia apenas na etapa de aprendizado, estabelecendo dependências entre o ajuste dos pesos de um neurônio e os pesos de seus vizinhos, segundo uma determinada função de vizinhança, como se mostra na descrição do processo de aprendizado.

No processo de aprendizado, áreas específicas de neurônios vão sendo ativadas. Ao final do processo, a rede SOM fica dividida em áreas especialistas, responsáveis pela classificação de padrões específicos, semelhante ao cérebro, onde cada região é responsável por uma atividade específica.

Os mapas auto-organizados utilizam neurônios suja saída é a que segue:
\begin{equation} \label{eq_neuronio_kohonen}
    y_i=e^{-||\mathbf{x}-\mathbf{w}_i||^2},
\end{equation}
onde $\mathbf{w}_i=(w_{i,1},w_{i,2},\dots,w_{i,n})^T$ é o vetor de pesos do $i$-ésimo neurônio, para $1\leq i\leq m$ e $1\leq j\leq n$.

Os pesos das redes SOM podem ser ajustados pelo seguinte procedimento modificado, baseado no procedimento clássico de ajuste de redes SOM, mas tendo como critérios de parada o número máximo de iterações e a soma dos ajustes dos pesos:
\begin{enumerate}
    \item Inicializar os pesos $w_{i,j}(0)$, onde $1\leq i\leq m$ e $1\leq j\leq n$, com valores aleatórios e ne\-ces\-sa\-ri\-a\-men\-te diferentes.
    \item Entrar com os vetores posição $\mathbf{r}_i$, correspondentes à posição do i-ésimo neurônio na grade, onde $1\leq i\leq m$.
    \item Inicializar a taxa de aprendizado inicial $\eta_0$, onde $0<\eta_0<1$.
    \item Inicializar a largura topológica $\sigma_0$. A largura topológica é a largura da função de vizinhança gaussiana $h_{i,k}(t)$, utilizada para definir a vizinhança do neurônio vencedor a ser treinada.
    \item Repetir até $\delta(t)=0$ ou um máximo de $N_{\max}$ iterações:
    \begin{enumerate}
        \item Para cada padrão $\mathbf{x}=(x_1, x_2, \dots, x_n)^T$ do conjunto de entrada $\Psi=\{\mathbf{x}^{(l)}\}_{l=1}^L$, repetir:
        \begin{enumerate}
            \item Calcular as saídas $y_i(t)$ dos neurônios, onde $1\leq i\leq m$.
            \item Calcular $y_{\max}(t)$:
            $$ y_{\max}(t)=\max(y_1(t), y_2(t), \dots, y_m(t)). $$
            \item Calcular o índice $k$ do neurônio vencedor:
            $$ y_i=y_{\max}\Rightarrow k=i. $$
            \item Calcular o ajuste $\Delta w_{i,j}(t)$:
            $$
            \Delta w_{i,j}(t)=\eta(t)h_{i,k}(t)(x_j(t)-w_{i,j}(t)),
            $$
            onde $1\leq i\leq m$ e $1\leq j\leq n$.
            \item Atualizar os pesos:
            $$ w_{i,j}(t+1)=w_{i,j}(t)+\Delta w_{i,j}(t), $$
            onde:
            $$ \sigma(t)=\sigma_0 \exp\left(-\frac{t}{\tau_1}\right), $$
            $$ h_{i,k}(t)=\exp\left(-\frac{d_{i,k}^2}{2\sigma^2(t)}\right), $$
            $$ \eta(t)=\eta_0 \exp\left(-\frac{t}{\tau_2}\right), $$
            onde $\tau_1$ e $\tau_2$ são \emph{constantes de tempo} e:
            $$ d_{i,k}=||\mathbf{r}_i-\mathbf{r}_k||, $$
            para $1\leq i\leq m$ e $1\leq j\leq n$.
        \end{enumerate}
        \item Calcular $\delta(t)$ da forma que segue:
        $$ \delta(t)=\sum_{i=1}^m \sum_{j=1}^n |\Delta w_{i,j}(t)|. $$
    \end{enumerate}
\end{enumerate}

As redes SOM podem ser utilizadas também como etapa de classificação ou agrupamento de dados para uma RNA supervisionada \cite{ham2001,kovacs1996,braga2000}, tendo diversas aplicações potenciais em reconhecimento de padrões na Engenharia Biomédica como, por exemplo, o auxílio ao diagnóstico da epilepsia a partir de sintomas \cite{sala2004}.

\subsection{Classificador LVQ de Kohonen}

O \emph{algoritmo LVQ}, ou algoritmo de quantização vetorial por aprendizagem (LVQ, \emph{Learning Vector Quantization}) consiste em uma técnica supervisionada de treinamento que tanto pode ser utilizada para adicionar novos padrões a uma rede SOM já treinada quanto permite criar uma rede neural supervisionada com a mesma arquitetura de uma rede SOM \cite{braga2000,haykin2001}.

Dado o conjunto de treinamento $\Psi=\{\mathbf{x}^{(l)},C^{(l)}\}_{l=1}^L$, onde $C^{(l)}$ é a classe correta \emph{a priori}, calcula-se a saída da rede para cada elemento $\mathbf{x}$ de $\Psi$ pertencente à classe $C$ e, caso o neurônio vencedor seja o $i$-ésimo, considerando que este neurônio está associado à classe $C_i$, para uma rede de $n$ entradas e $m$ saídas projetada para reconhecer padrões de um universo $\Omega=\{C_1, C_2, \dots, C_m\}$, onde $1\leq i\leq m$, os pesos serão atualizados da forma que segue \cite{haykin2001,braga2000}:
\begin{equation}
  w_{i,j}(t+1)=\left\{
  \begin{array} {ll}
      {w_{i,j}(t)+\eta(t)(x_j(t)-w_{i,j}(t)),} & {C_i=C}\\
      {w_{i,j}(t)-\eta(t)(x_j(t)-w_{i,j}(t)),} & {C_i\neq C}
    \end{array}
  \right.,
\end{equation}
onde $1\leq j\leq n$ e $\eta$ é a taxa de aprendizado.

A inicialização da rede LVQ é idêntica à da rede SOM. A rede LVQ pode usar como critério de parada tanto o número de iterações quanto o erro de classificação, dado que o treinamento é supervisionado.

\subsection{Morfologia Matemática e Espectro de Padrões}

Uma \emph{imagem} $f:S\rightarrow K$ é definida como sendo a função que mapeia um vetor bidimensional $u\in S$ no ponto $f(u)\in K$, onde $S$ é a \emph{grade} ou \emph{suporte} da imagem, $u$ é o \emph{pixel} da grade $S$; $f(u)\in K$ é o \emph{valor} do \emph{pixel} $u\in S$.

De forma mais rigorosa, \cite{candeias1997} define \emph{pixel} como sendo o par $(u,f(u))$, ou seja, o vetor formado pela sua posição $u$ na grade $S$ e seu valor associado $f(u)$. Já $K$ é o conjunto dos valores possíveis de $f(u)$. Denota-se $K^S$ o \emph{conjunto de todas as imagens possíveis} na grade $S$ em $K$ \cite{candeias1997}.

Quando a grade $S$ é \emph{discreta}, isto é, $S\subseteq \mathbb{Z}^2$, diz-se que $f:S\rightarrow K$ é uma \emph{imagem discreta}. $K$ é definido como sendo da forma $K=V^p$, onde $V\in \mathbb{R}$ é o \emph{conjunto dos níveis de cinza} e $p\in \mathbb{N}^*$ é o \emph{número de bandas} presentes na imagem $f$.

Assim, seja $f$ uma imagem de $p$ bandas $f:S\rightarrow V^p$, $f$ pode ser denotada de forma equivalente por
$$
f(u)=\{f_1(u),f_2(u),\dots,f_p(u)\},~~u\in S,
$$
onde $f_j(u)$ é a $j$-ésima banda de $f:S\rightarrow V^p$, para $f_j:S\rightarrow V$ e $1\leq j\leq p$.

Quando $p=1$ diz-se que a imagem é uma imagem em \emph{níveis de cinza} ou \emph{monocromática}. Uma imagem é dita ser uma \emph{imagem colorida} quando $p=3$ e \emph{multiespectral} quando $4\leq p\leq 100$ \cite{landgrebe2002}. Já uma imagem é dita \emph{hiperespectral} quando $p>100$ \cite{landgrebe2002}. Para generalizar, pode-se dizer que uma imagem é multiespectral quando $p>1$.

Uma imagem $f:S\rightarrow K$ é dita ser uma \emph{imagem digital} quando, além de ser discreta, ela é \emph{quantizada}, ou seja, seus \emph{pixels} assumem apenas vetores discretos. Matematicamente, $f:S\rightarrow K$ é uma imagem digital quando $K=\{0,1,\dots,k\}^p$ e $S\in \mathbb{Z}$, onde $p\in \mathbb{N}^*$.

Quando uma imagem digital é representada com $N$ bits, $k=2^N-1$. Assim, uma imagem digital $f:S\rightarrow K$ de $8$ bits tem $K=\{0,1,\dots,255\}^p$, enquanto que imagens digitais de $16$ bits tem $K=\{0,1,\dots,65535\}^p$. Conseqüentemente, uma \emph{imagem binária de banda única} $f:S\rightarrow K$ é uma imagem digital com $K=\{0,1\}$.

Por uma questão de conveniência, é comum representar as imagens digitais como imagens discretas normalizadas, substituindo $K=\{0,1,\dots,k\}^p$ por $\overline{K}=[0,1]^p$. Desta forma não é preciso explicitar quantos bits são utilizados na representação dos valores possíveis dos \emph{pixels}. Assim, uma imagem $f:S\rightarrow \{0,1,\dots,k\}^p$ passa a ser representada por $\tilde{f}:S\rightarrow [0,1]^p$.

É importante realçar que as imagens binárias são apenas casos par\-ti\-cu\-la\-res das imagens em níveis de cinza. Assim, uma imagem binária $f:S\rightarrow \{0,k\}^p$ passa a ser denotada utilizando a notação normalizada por $\tilde{f}:S\rightarrow \{0,1\}^p$.

Salvo indicação em contrário, neste texto uma imagem $f:S\rightarrow K$ sempre será representada por sua versão normalizada, sendo $K^S$ o conjunto de todas as imagens normalizadas, onde $K=[0,1]^p$, para $p$ bandas.

Usualmente $S$ é uma região \emph{retangular} de $\mathbb{Z}^2$, podendo-se representar as imagens digitais em nível de cinza $f:S\rightarrow \{0,1,\dots,k\}$ como \emph{matrizes} \cite{klette1995,dougherty1987}. \cite{dougherty1987} representaria uma imagem $f:S\rightarrow V$ da seguinte forma, para $S$ sendo uma grade $m\times n$ centrada em $(r,s)$:
$$
f=\left[
\begin{array}{cccc}
a_{11} & a_{12} & \cdots & a_{1n}\\
a_{21} & a_{22} & \cdots & a_{2n}\\
\vdots & \vdots & \ddots & \vdots\\
a_{m1} & a_{m2} & \cdots & a_{mn}
\end{array}
\right]_{(r,s)},
$$
onde $a_{ij}\in V$, para $1\leq i\leq m$ e $1\leq j\leq n$. Isto implica dizer que, segundo esta notação: $$ f(u)=a_{ij}\Rightarrow u=(i-r,j-s),~~u\in S. $$

Note que, neste caso, a grade $S$ é uma região retangular de $\mathbb{Z}^2$ do tipo:
$$ S= \{(1-r),(2-r),\dots,(m-r)\} \times \{(1-s),(2-s),\dots,(n-s)\}. $$

Esta será a notação utilizada ao longo deste texto sempre que for conveniente representar uma imagem como uma matriz.

As operações básicas entre imagens em Morfologia Matemática são a \emph{comparação}, a \emph{subtração}, a \emph{intersecção} e a \emph{união}.

É importante afirmar que as operações básicas da Morfologia Matemática só podem ser aplicadas a imagens que mantenham uma relação de ordem entre os \emph{pixels}. Somente as imagens em níveis de cinza e as imagens binárias mantêm tal relação. No caso das imagens multiespectrais ou hiperespectrais, é necessário trabalhar banda a banda ou supor algum artifício para manter a relação de ordem. A relação de ordem é a base da Morfologia Matemática \cite{candeias1997}.

Dadas duas imagens $f_1:S\rightarrow [0,1]$ e $f_2:S\rightarrow [0,1]$, a \emph{comparação} $\leqq$ é a operação de $[0,1]^S\times [0,1]^S$ em $[0,1]^S$ definida como segue \cite{candeias1997}:
\begin{equation}\label{mmeq1}
(f_1\leqq f_2)(u)=\left\{
\begin{array}{ll}
1, & {f_1(u)\leq f_2(u)}\\
0, & c.c.
\end{array}
\right.,~~\forall u\in S.
\end{equation}

Dada uma imagem $f:S\rightarrow [0,1]$, o \emph{negativo} de $f$, denotado por $\overline{f}$ ou $\sim f$, é o operador de $[0,1]^S$ em $[0,1]^S$ definido como sendo \cite{banon1994}:
\begin{equation}\label{mmeq2}
\overline{f}(u)=1-f(u),~~\forall u\in S.
\end{equation}

A \emph{união} entre $f_1:S\rightarrow [0,1]$ e $f_2:S\rightarrow [0,1]$ é o operador de $[0,1]^S\times [0,1]^S$ em $[0,1]^S$, denotado $f_1\vee f_2$, definido como segue \cite{candeias1997}:
\begin{equation}\label{mmeq3}
(f_1\vee f_2)(u)=\max\{f_1(u),f_2(u)\},~~\forall u\in S.
\end{equation}

A \emph{intersecção} entre $f_1:S\rightarrow [0,1]$ e $f_2:S\rightarrow [0,1]$ é o operador de $[0,1]^S\times [0,1]^S$ em $[0,1]^S$, denotado $f_1\wedge f_2$, definido como \cite{candeias1997}:
\begin{equation}\label{mmeq4}
(f_1\wedge f_2)(u)=\min\{f_1(u),f_2(u)\},~~\forall u\in S.
\end{equation}

Pode-se notar que as operações de união e intersecção entre imagens se assemelham às operações de mesmo nome entre conjuntos nebulosos. A semelhança entre estas operações da Morfologia Matemática e da Lógica Nebulosa cresce com a adoção da notação de imagem normalizada, o que faz com que, de certa forma, as imagens normalizadas se assemelhem a conjuntos nebulosos.

A \emph{subtração} ou \emph{diferença} entre $f_1:S\rightarrow [0,1]$ e $f_2:S\rightarrow [0,1]$ é operação de $[0,1]^S\times [0,1]^S$ em $[0,1]^S$ dada por \cite{banon1994}:
\begin{equation}\label{mmeq5}
(f_1\sim f_2)(u)=(f_1\wedge \overline{f}_2)(u),~~\forall u\in S.
\end{equation}

A partir destas operações é possível definir os \emph{operadores básicos} da Morfologia Matemática: a \emph{dilatação}, a \emph{erosão}, o \emph{gradiente morfológico}, a \emph{abertura}, o \emph{fechamento}, as \emph{aberturas e fechamentos por reconstrução} e as \emph{dilatações e erosões condicionais}.

Pode-se dizer que a Morfologia Matemática é uma teoria \emph{construtiva}, devido ao fato de que operadores mais complexos podem ser construídos a partir dos mais simples. Destes, os operadores fundamentais são a \emph{dilatação} e a \emph{erosão}, originando-se todos os outros a partir de combinações destes \cite{soille2004}.

A \emph{dilatação} de $f:S\rightarrow [0,1]$ por $g:S\rightarrow [0,1]$, denotada $\delta_g(f)$ ou $f\oplus g$ (soma de Minkowsky, no caso binário \cite{candeias1997}), é um operador de $[0,1]^S\times [0,1]^S$ em $[0,1]^S$ definido como:
\begin{equation}\label{mmeq6}
\delta_g(f)(u)=(f\oplus g)(u):=\bigvee_{v\in S}f(v)\wedge
g(u-v),~~\forall u\in S,
\end{equation}
onde $g$ é denominado \emph{elemento estruturante}. Na prática, a dilatação transforma a imagem original de forma a fazer o elemento estruturante se ``encaixar'' na imagem, ou seja, a imagem original $f$ será modificada de modo a fazer as áreas semelhantes a $g$ ``crescerem'', ``encaixando'' $g$.

Visualmente, a dilatação resulta no \emph{crescimento das áreas claras} e na \emph{eliminação das áreas escuras menores do que o elemento estruturante}, considerando o fato de que, usualmente, os \emph{pixels} mais claros são representados por valores mais próximos de $1$, enquanto os mais escuros, por valores mais próximos de $0$.

A \emph{erosão} de $f:S\rightarrow [0,1]$ por $g:S\rightarrow [0,1]$, denotada $\epsilon_g(f)$ ou $f\ominus g$ (subtração de Minkowsky, no caso binário \cite{candeias1997}), é um operador de $[0,1]^S\times [0,1]^S$ em $[0,1]^S$ definido como:
\begin{equation}\label{mmeq7}
\epsilon_g(f)(u)=(f\ominus g)(u):=\bigwedge_{v\in S}f(v)\vee
\overline{g}(v-u),~~\forall u\in S.
\end{equation}

Na prática, a erosão transforma a imagem original $f$ de modo a fazer as áreas semelhantes a $g$ ``diminuirem'', ``encaixando'' $\overline{g}$, conforme o exemplo anterior.

Visualmente, a erosão resulta no \emph{crescimento das áreas escuras} e na \emph{eliminação das áreas claras menores do que o elemento estruturante}, considerando o fato de que, usualmente, os \emph{pixels} mais claros são representados por valores mais próximos de $1$, enquanto os mais escuros, mais próximos de $0$.

Em \cite{banon1994} a dilatação e a erosão são mostradas esquematicamente por meio de exemplos e ilustrações, que mostram o elemento estruturante percorrendo uma ima\-gem binária. É uma abordagem bastante ilustrativa, que dá ao leitor uma idéia de como são implementadas de forma algorítmica a dilatação e a erosão.

A Tabela \ref{mmtabcomp} mostra as expressões das operações de \emph{dilatação}, \emph{erosão} e \emph{convolução}, po\-den\-do-se perceber a semelhança entre estas três operações.
\begin{table}
  \centering
  \caption{Operações de dilatação, erosão e convolução}\label{mmtabcomp}
\begin{tabular}{|l|l|}
\hline
\emph{Operação} & \emph{Expressão}\\
\hline \hline Dilatação & {$(f\oplus g)(u)=\bigvee_{v\in
S}f(v)\wedge g(u-v)$}\\
Erosão & {$(f\ominus g)(u)=\bigwedge_{v\in
S}f(v)\vee \overline{g}(v-u)$}\\
Convolução & {$(f*g)(u)=\sum_{v\in S}f(v)g(u-v)$}\\
\hline
\end{tabular}
\end{table}

Quanto às suas propriedades, a \emph{dilatação} é uma \emph{operação extensiva} devido ao fato de $\delta_g(f)(u)\geq f(u),~\forall u\in S$, enquanto que a \emph{erosão} é uma operação \emph{anti-extensiva}, pois $\epsilon_g(f)(u)\leq f(u),~\forall u\in S$ \cite{ulisses1994}.

A \emph{anti-dilatação} de $f:S\rightarrow [0,1]$ por $g:S\rightarrow [0,1]$, denotada por $\delta_g^\mathrm{a}(f)$, é dada por:
\begin{equation}\label{mmeq11}
\delta_g^\mathrm{a}(f)(u):=\overline{\delta_g(f)}(u),~~\forall
u\in S.
\end{equation}

A \emph{anti-erosão} de $f:S\rightarrow [0,1]$ por $g:S\rightarrow [0,1]$, denotada por $\epsilon_g^\mathrm{a}(f)$, é dada por:
\begin{equation}\label{mmeq12}
\epsilon_g^\mathrm{a}(f)(u):=\overline{\epsilon_g(f)}(u),~~\forall
u\in S.
\end{equation}

A \emph{$n$-dilatação} de $f:S\rightarrow [0,1]$ por $g:S\rightarrow [0,1]$, denotada por $\delta_g^n(f)$, é dada por, para $n>1$:
\begin{equation}\label{mmeq13}
\delta_g^n(f)(u):=\underbrace{\delta_g\delta_g\cdots\delta_g}_n(f)(u),~~\forall
u\in S.
\end{equation}

De forma semelhante, a \emph{$n$-erosão} de $f:S\rightarrow [0,1]$ por $g:S\rightarrow [0,1]$, denotada por $\epsilon_g^n(f)$, é dada por, para $n>1$:
\begin{equation}\label{mmeq14}
\epsilon_g^n(f)(u):=\underbrace{\epsilon_g\epsilon_g\cdots\epsilon_g}_n(f)(u),~~\forall
u\in S.
\end{equation}

O elemento estruturante \emph{quadrado $3\times 3$} é dado por:
$$
b=\left[
\begin{array}{lll}
1 & 1 & 1\\
1 & 1 & 1\\
1 & 1 & 1
\end{array}
\right]_{(2,2)}.
$$

O elemento estruturante \emph{cruz $3\times 3$} é dado por:
$$
b=\left[
\begin{array}{lll}
0 & 1 & 0\\
1 & 1 & 1\\
0 & 1 & 0
\end{array}
\right]_{(2,2)}.
$$

Em \cite{banon1994} é demonstrado que qualquer elemento estruturante de dimensões maiores que $3\times 3$ pode ser decomposto em vários $3\times 3$. Isto quer dizer que uma dilatação por um elemento estruturante $h$ de dimensões $m\times n$, tal que $m>3$ e $n>3$, pode ser executada por meio de um certo número de dilatações por $b_k$ de dimensões $3\times 3$, onde cada $b_k$ é um elemento estruturante que constitui $h$ \cite{banon1994}.

Uma das aplicações mais importantes da Morfologia Matemática é a implementação de \emph{filtros morfológicos}, os quais consistem em filtros não lineares baseados em operações morfológicas. Os filtros morfológicos mais comuns são a \emph{abertura}, o \emph{fechamento}, a \emph{$n$-abertura}, o \emph{$n$-fechamento} e os \emph{filtros alternados}.

Segundo \cite{candeias1997}, uma transformação $\xi$ é um \emph{filtro morfológico} se for ao mesmo tempo \emph{idempotente} e \emph{crescente}.

A transformação $\xi$ é \emph{crescente} se, dadas duas imagens $f:S\rightarrow K$ e $g:S\rightarrow K$, a seguinte condição é satisfeita \cite{candeias1997}:
\begin{equation}\label{mmeq20}
f(u)\leq g(u) \Rightarrow \xi(f)(u)\leq \xi(g)(u),~~\forall u\in S.
\end{equation}

Por sua vez, a transformação $\xi$ é \emph{idempotente} se \cite{candeias1997}:
\begin{equation}\label{mmeq21}
\xi\xi(f)(u)=\xi(f)(u),~~\forall u\in S,
\end{equation}
ou seja, se $\xi$ é idempotente, a imagem $f$ \emph{não é alterada} por aplicações sucessivas da mesma transformação $\xi$.

O filtro morfológico de \emph{abertura} é utilizado para ``eliminar'' regiões claras na ima\-gem que não possam estar contidas no elemento estruturante. Assim, a abertura realiza na imagem uma espécie de ``peneiramento'' dos pontos claros. Como resultado, fazendo uma analogia entre uma imagem em níveis de cinza e um relevo topográfico, os picos (pontos de máximo ou regiões mais claras) inferiores em tamanho ao elemento estruturante são ``eliminados''.

Matematicamente, a \emph{abertura} de $f:S\rightarrow [0,1]$ pelo elemento estruturante $g:S\rightarrow [0,1]$, denotada $\gamma_g(f)$, é a transformação de $[0,1]^S$ em $[0,1]^S$ definida por \cite{candeias1997}:
\begin{equation}\label{mmeq22}
\gamma_g(f):=\delta_g\epsilon_g(f).
\end{equation}

A abertura $\gamma_g(f)$ é uma transformação \emph{anti-extensiva}, pois:
\begin{equation}\label{mmeq23}
\gamma_g(f)(u)\leq f(u),~~\forall u\in S.
\end{equation}

O filtro morfológico de \emph{fechamento} é utilizado para ``eliminar'' regiões escuras na imagem que não possam estar contidas no elemento estruturante. Assim, o fechamento realiza na imagem uma espécie de ``peneiramento'' dos pontos escuros. Como resultado, novamente utilizando uma analogia entre uma imagem em níveis de cinza e um relevo topográfico, os vales (pontos de mínimo ou regiões mais escuras) inferiores em tamanho ao elemento estruturante são ``eliminados''.

Matematicamente, o \emph{fechamento} de $f:S\rightarrow [0,1]$ pelo elemento estruturante $g:S\rightarrow [0,1]$, denotado $\phi_g(f)$, é a transformação de $[0,1]^S$ em $[0,1]^S$ definida da forma que segue \cite{candeias1997}:
\begin{equation}\label{mmeq24}
\phi_g(f):=\epsilon_g\delta_g(f).
\end{equation}

O fechamento $\phi_g(f)$ é uma transformação \emph{extensiva}, pois:
\begin{equation}\label{mmeq25}
f(u)\leq \phi_g(f)(u),~~\forall u\in S.
\end{equation}

A \emph{$n$-abertura} de $f:S\rightarrow [0,1]$ por $g:S\rightarrow [0,1]$, denotada $\gamma_g^n(f)$, é definida da forma que segue \cite{candeias1997}:
\begin{equation}\label{mmeq26}
\gamma_g^n(f):=\delta_g^n\epsilon_g^n(f).
\end{equation}

O \emph{$n$-fechamento} de $f:S\rightarrow [0,1]$ por $g:S\rightarrow [0,1]$, denotado $\phi_g^n(f)$, é definido como segue \cite{candeias1997}:
\begin{equation}\label{mmeq27}
\phi_g^n(f):=\epsilon_g^n\delta_g^n(f).
\end{equation}

Dentre as ferramentas da Morfologia Matemática para a descrição de tamanho e forma em análise de imagens, as \emph{granulometrias} cons\-ti\-tuem algumas das mais utilizadas. A idéia básica de uma granulometria é realizar diversas \emph{crivagens} (ou \emph{peneiramentos}) seguidas de medições quantitativas dos resíduos, tais como medição de área ou perímetro. Ou seja, realiza-se um \emph{peneiramento} da ima\-gem original utilizando uma certa \emph{peneira} e depois medem-se os \emph{grãos} que passaram; em seguida, utiliza-se uma \emph{peneira} com \emph{furos menores} e novamente se medem os resíduos, e assim por diante. No caso, o \emph{tamanho dos furos da peneira} é determinado pela dimensão do elemento estruturante utilizado na transformação granulométrica.

A família de transformações $\{\psi_j\}$, parametrizadas com um parâmetro $j\geq 0$, onde
$$
\psi_0(f)(u)=f(u),~\forall u\in S,~f:S\rightarrow [0,1],
$$
é chamada de uma \emph{granulometria} se constituir um \emph{critério de forma} \cite{ulisses1994}, ou seja:
\begin{eqnarray}\label{mmeq32}
\psi_j(f)(u)\leq f(u),\\
f(u)\leq g(u)\Rightarrow\psi_j(f)(u)\leq\psi_j(g)(u),\\
\psi_j[\psi_k(f)](u)=\psi_k[\psi_j(f)](u)=\psi_{\max(j,k)}(f)(u),~~\forall
j,k\geq 0,
\end{eqnarray}
$\forall u\in S$, onde $g:S\rightarrow [0,1]$.

A última condição expressa a idéia intuitiva de que dois \emph{peneiramentos} seguidos são equivalentes a um só \emph{peneiramento} com a \emph{peneira de buracos menores} ($\max(j,k)$), já que o índice $j$ da transformação $\psi_j$ expressa o rigor do \emph{peneiramento}: quanto maior o $j$, mais rigoroso o \emph{peneiramento}, ou seja, menores os \emph{grãos} passantes.

É possível demonstrar que o conjunto de transformações morfológicas que constitui um \emph{critério de forma} é o \emph{conjunto das $j$-aberturas} por um elemento estruturante \emph{disco digital} \cite{ulisses1994}. Os discos digitais mais elementares são o quadrado $3\times 3$ e a cruz $3\times 3$ \cite{ulisses1994}. Assim, $\{\gamma_g^j\}$ é uma \emph{granulometria}, onde $j\geq 0$ e $g$ é um disco digital.

Logo, uma granulometria passa a ser expressa da forma $\{\psi_j=\gamma_g^j\},~r\geq 0$, onde $g$ é um disco digital \cite{ulisses1994}.

Analogamente, define-se uma \emph{anti-granulometria} como sendo o conjunto dos $j$-fe\-cha\-men\-tos $\{\phi_g^j\}$, onde $j\geq 0$ e $g$ é um disco digital \cite{ulisses1994}.

Seja uma imagem $f:S\rightarrow [0,1]$. A função $V:\mathbb{Z}_+\rightarrow \mathbb{R}$ é definida como sendo \cite{ulisses1994}:
\begin{equation} \label{mmeq33}
V(k)=\sum_{u\in S}\gamma_g^k(f)(u),
\end{equation}
onde $g$ é a cruz ou o quadrado $3\times 3$. Assim, a função $\Xi:\mathbb{Z}_+\rightarrow [0,1]$ é definida da forma que segue:
\begin{equation} \label{mmeq34}
\Xi[k]=1-\frac{V(k)}{V(0)},~~k\geq 0.
\end{equation}

Considerando a imagem $f$ um \emph{conjunto aleatório} e observando que $\Xi$ é uma função monotônica, crescente e limitada em $[0,1]$, pois $V(k+1)<V(k),~\forall k\geq 0$, podemos afirmar que $\Xi$ é uma \emph{função de distribuição acumulada discreta} associada à imagem $f$ \cite{ulisses1994}. Assim, pode-se definir uma \emph{função densidade discreta} $\xi:\mathbb{Z}_+\rightarrow \mathbb{R}_+$ por meio da diferença:
\begin{equation}\label{mmeq35}
\xi[k]=\Xi[k+1]-\Xi[k],~~k\geq 0.
\end{equation}

A função densidade definida na equação \ref{mmeq35} é de\-no\-mi\-na\-da \emph{espectro de padrões} \cite{ulisses1994}. O espectro de padrões, ou \emph{espectro morfológico}, é uma espécie de \emph{histograma de tamanho e forma}, sendo \emph{único} para uma imagem específica $f$ e um determinado elemento estruturante $g$, desde que $f$ seja \emph{binária} \cite{ulisses1994}.

Uma vez que as transformações granulométricas têm a propriedade de ``espalhar'' informações de tamanho e forma nas imagens resíduo, muitas outras representações diferentes do histograma de padrões dos objetos de interesse podem ser geradas, desde que uma quantidade adequada de transformações seja utilizada. Como é de se esperar, o número de transformações (ou \emph{peneiramentos}) necessário, ou seja, o maior valor de $k$, varia de acordo com a
aplicação, pois está diretamente associado à dimensão do vetor de atributos a ser utilizado na aplicação de reconhecimento de padrões.

\subsection{Triagem virtual}

A Triagem Virtual consiste em agrupar imagens médicas semelhantes para facilitar a sua distribuição para os especialistas responsáveis pela análise de imagens e, por conseguinte, pelo diagnóstico. Neste trabalho foi estudada uma aplicação baseada em imagens de imuno-histoquímica da placenta e do pulmão do tipo CD68 \cite{santos2003_2,santos2004}. Essas imagens são utilizadas para se poder auxiliar o patologista no seu diagnóstico fazendo comparação com outros pacientes e seus respectivos diagnósticos. Fazendo a comparação visual ou com o auxílio de algum programa o patologista pode analisar o caso atual de maneira a reduzir a probabilidade de erro.

A busca de imagens em uma base de imagens médicas pode ser prejudicada por fatores como os que seguem:
\begin{enumerate}
  \item As imagens foram agrupadas de maneira subjetiva: influência da subjetividade de cada patologista na análise pode gerar inconsistências no agrupamento;
  \item A recuperação das imagens pode retornar um número insuficiente de imagens ou um conjunto de imagens com pouca relevância com o problema;
  \item O sistema pode demorar muito para recuperar imagens relevantes: mesmo que o sistema recupere uma quantidade grande de imagens relevantes, a quantidade de tempo gasta com essa recuperação também é importante para o usuário.
\end{enumerate}

\begin{figure}
	\centering
		\includegraphics[width=0.85\textwidth]{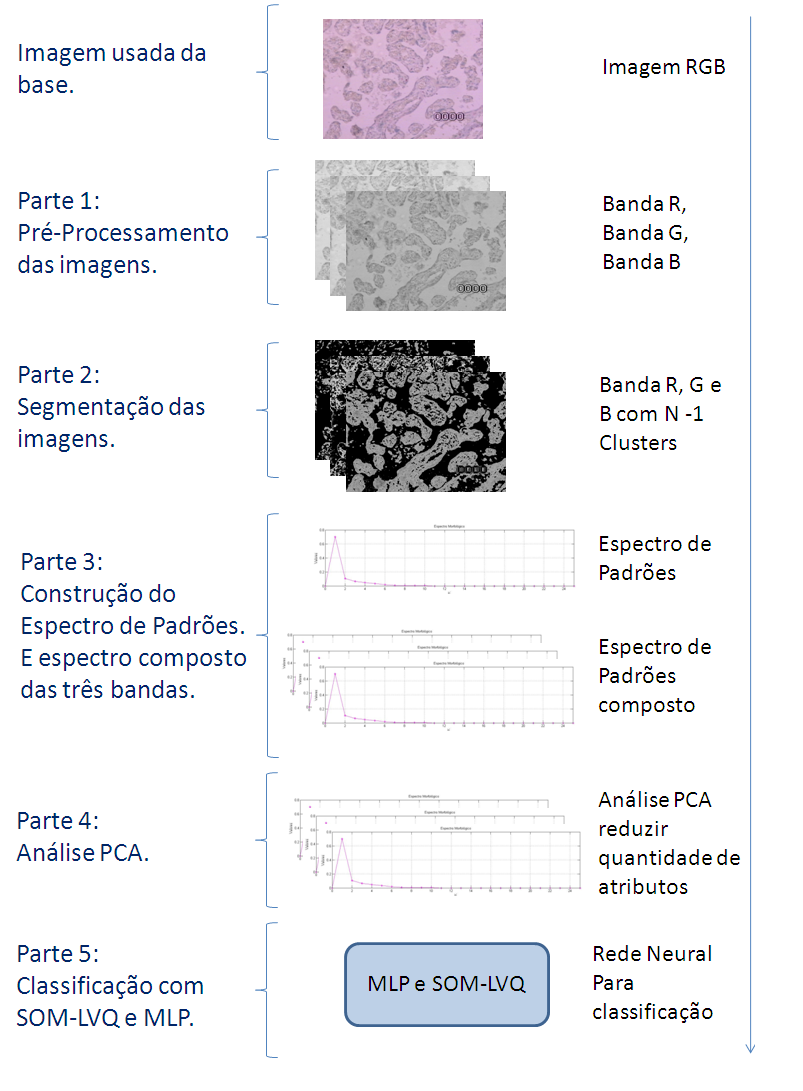}
	\caption{Metodologia empregada: etapas da classificação para triagem virtual}
	\label{fig:metodologia}
\end{figure}

Neste trabalho as imagens são representadas por meio de vetores de atributos construídos usando uma concatenação dos espectros de padrões das bandas R, G e B que compõem as imagens coloridas; em seguida são feitos estudos usando PCA para redução da dimensionalidade do vetor de atributos, objetivando a redução do custo computacional na tarefa de classificação. Assim, a metodologia de busca da solução do problema de triagem virtual estudado neste trabalho foi dividida em cinco etapas: (1) pré-processamento, (2) segmentação das imagens, (3) construção dos espectros de padrões, (4) análise PCA sobre os espectros e (5) uso de SOM-LVQ e MLP para classificar os vetores de características. Essas etapas são ilustradas no esquema da Figura \ref{fig:metodologia}. Na última fase foram utilizados conjuntos de treino e teste para poder avaliar quantitativamente a taxa de acertos na triagem das imagens.

A segmentação das imagens R, G e B foi feita com uma implementação de k-médias com 4 classes. As imagens utilizadas foram analisadas e na média existiam de 3 a 6 classes com o maior número de \emph{pixels}. Dentre essas classes, apenas 4 seriam suficientes para representar as imagens. Com isso, foram escolhidas 3 classes para representar os objetos presentes na imagem, enquanto uma quarta classe representa fundo da imagem. Ao fim da segmentação a quarta classe é removida, restando apenas três classes.

Foram calculados os espectros de padrões para as imagens R, G e B usando 25 iterações para compor cada espectro, ficando assim cada banda espectral com um espectro de 25 atributos. Em seguida esses espectros foram concatenados em um vetor único, obtendo-se ao fim um vetor de 75 dimensões.

A análise PCA foi o passo que teve como entrada a matriz unificada gerada, que representava as imagens dos dois tipos. Essa matriz foi processada para obter a menor representação, i.e. com a menor quantidade de atributos, mas mantendo uma boa representação dos dados.

Foi usada uma rede SOM-LVQ e uma rede MLP para classificar os dados usando a validação cruzada como condição de parada da rede. As duas redes foram testadas usando a matriz resultante da análise PCA e a matriz sem o uso da análise PCA. A classificação foi dividida em duas etapas, como descrito adiante.

Primeiro foi feita uma análise para se descobrir a melhor configuração de rede para o problema. Para isso foi estabelecida uma bateria de testes que fez os treinamentos da rede variando a quantidade de neurônios na camada escondida. Foram utilizados de 10 a 30 neurônios com 20 repetições para cada rede. Em cada passo foi guardada a melhor rede, baseando-se na taxa de acertos e na média de acertos de todas as redes. O segundo passo foi repetir os testes com a melhor rede e obter a média, desvio padrão e mediana da taxa de acertos.

A matriz de entrada foi dividida em 19\% para teste e 81\% para treino e os valores foram embaralhados no início dos treinos. Com isso, o resultado obtido foi uma comparação da abordagem usando o PCA, para reduzir a complexidade, mas com o intuito de manter uma boa representação das imagens com uma abordagem de referência usando todos os parâmetros do vetor de entrada.

\begin{figure}
  \centering
  \begin{minipage}[b]{0.48\linewidth}
    \centering
    \includegraphics[width=0.9\linewidth]{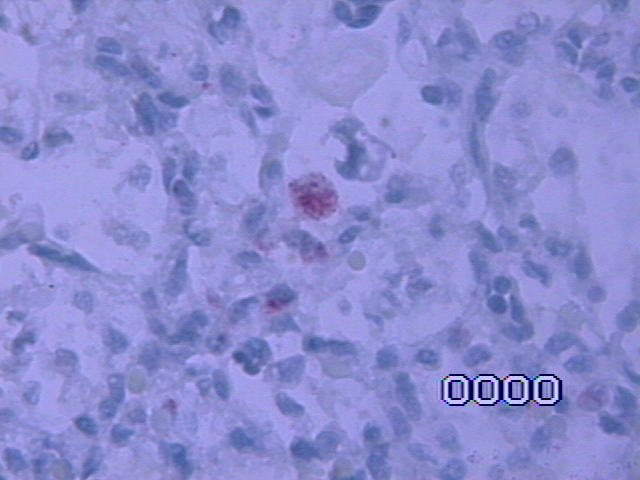}
    \\(a)\\
	  \includegraphics[width=0.9\linewidth]{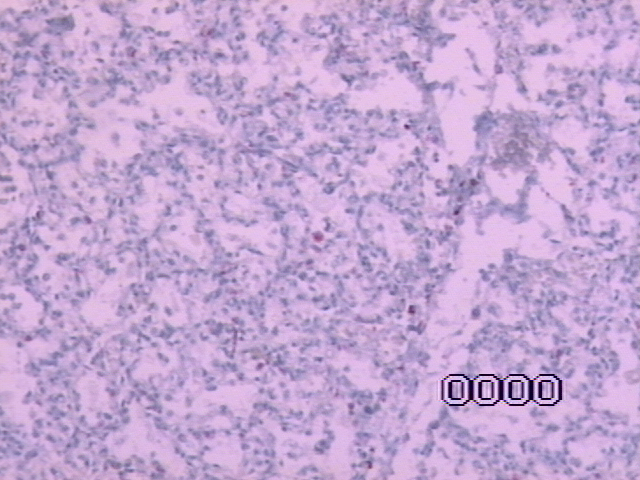}
    \\(b)\\
  \end{minipage}
  \begin{minipage}[b]{0.48\linewidth}
    \centering
    \includegraphics[width=0.9\linewidth]{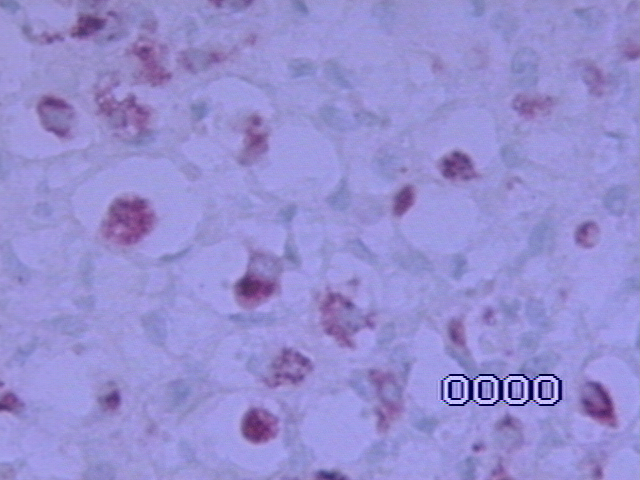}
    \\(c)\\
	  \includegraphics[width=0.9\linewidth]{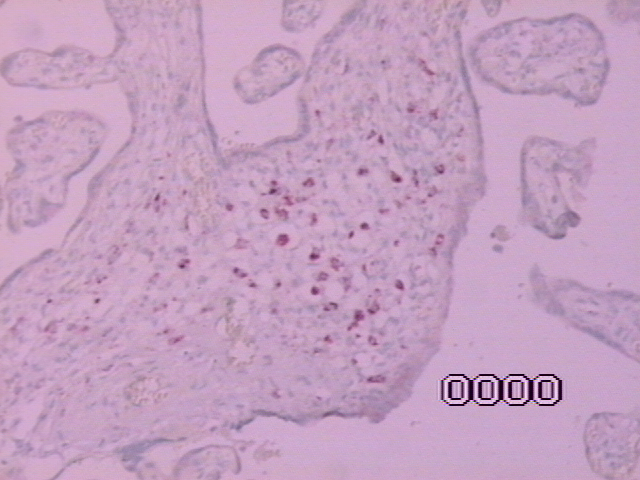}
    \\(d)\\
  \end{minipage}
  \caption{Tipos de imagens de imuno-histoquímica CD68: (a) NP com mais \emph{zoom}; (b) NP com menos \emph{zoom}; (c) P com mais \emph{zoom}; e (d) P com menos \emph{zoom}.}
  \label{fig:TiposImagens}
\end{figure}

Neste trabalho foram utilizadas imagens de imuno-histoquímica CD68 da placenta e do pulmão \cite{santos2003_2,santos2004}. As imagens foram diferenciadas pela nomenclatura P (pulmão) e NP (placenta). As imagens foram adquiridas usando dois tipos de \emph{zoom}. A Figura \ref{fig:TiposImagens} ilustra quatro de todas as imagens utilizadas. A base de dados tem um total de 113 imagens, sendo 53 imagens do tipo NP e 60 imagens do tipo P.

\section{Resultados Experimentais} \label{sec:quanti_resultados}

As imagens na aplicação de CBIR de Triagem Virtual foram representadas usando vetores multidimensionais baseados nos seguintes passos, e com os resultados que seguem:
\begin{enumerate}
  \item Cada imagem foi separada em bandas R, G e B;
  \item Cada banda foi segmentada, resultando uma imagem com quatro clusters, sendo o maior retirado;
  \item Foi construído o espectro de padrões da cada banda resultando em um vetor de 25 dimensões. Depois de serem unificados, resultaram em um vetor de 75 dimensões. As classes de imagens NP e P ficaram com matrizes 53 por 75 e 60 por 75, respectivamente. Por fim, foi gerada a matriz de características para representar a base de dimensão 113 por 75 pela união dessas duas matrizes;
  \item A análise PCA reduziu a complexidade da base para uma matriz 113 por 50. Obteve-se assim uma redução de 33,33\% na quantidade de atributos da matriz. Ou seja, uma redução de cerca de 1/3.
\end{enumerate}

Os testes de agrupamento das imagens foram feitos em duas etapas. Para implementar o método proposto para triagem virtual foi usado o ambiente Matlab na versão R2007b como plataforma para construção, teste e treinamento da metodologia proposta. Foram utilizados 22 padrões para teste e 91 padrões para treinamento. Desses valores de teste, 11 eram do tipo NP e 11 eram do tipo P. Dos 91 padrões para treinamento, 42 eram do tipo NP e 49 eram do tipo P. Todos os treinamentos usaram validação cruzada com parte dos dados de treinamento.

Após a primeira etapa de classificação foram geradas estatísticas para embasar a seleção da melhor rede. Os resultados são mostrados na Tabela \ref{tab:resultados1}, com a rede utilizada, a quantidade de atributos que foram usados como entrada da rede ($m$), a taxa de acerto da melhor rede ($\eta_{\mathrm{OTM}}$), a média ($\bar{\eta}$), o desvio padrão ($\Delta\eta$), a mediana ($\eta_{\mathrm{MED}}$) e a quantidade de neurônios da melhor rede ($n_{\mathrm{OTM}}$). 
\begin{table}[htbp]
	\centering
	\footnotesize
	\caption{Resultados da primeira etapa da classificação, usando PCA ($m=50$) e sem uso do PCA ($m=75$)}
		\begin{tabular} {c|c|c|c|c|c}
			{Rede} & {$m$} & {$\eta_{\mathrm{OTM}}$ (\%)} & {$\bar{\eta}\pm \Delta\eta$ (\%)} & {$\eta_{\mathrm{MED}}$ (\%)} & {$n_{\mathrm{OTM}}$}\\
			\hline
      {\multirow{2}{*}{MLP}} & {50} & {95,23} & {57,05 $\pm$ 0,48} & {68,18} & {22}\\
      {} & {75} & {90,47} & {53,83 $\pm$ 0,47} & {50,00} & {19}\\
      \hline
      {\multirow{2}{*}{SOM LVQ}} & {50} & {66,66} & {50,49 $\pm$ 0,23} & {40,91} & {10}\\
      {} & {75} & {80,95} & {47,14 $\pm$ 0,15} & {50,00} & {17}\\      
			\hline			
		\end{tabular}
	\label{tab:resultados1}
\end{table}

Nas Tabelas \ref{tab:resultados1} e \ref{tab:resultados2} são apresentados os resultados para MLP e SOM-LVQ, considerando quantidades de atributos 50 e 75, com e sem o uso de PCA, respectivamente. Para as redes MLP usando padrões obtidos com o uso de PCA ($m=50$) foi observada uma média de acertos de 57,05\% e um desvio padrão de 0,48\%; no entanto, foi observada uma mediana distante da média (mediana de 68,18\%), indicando que houve resultados díspares em relação à média. Isso fica evidente quando se observa a taxa de acerto da melhor rede, de 95,23\%. Neste caso, a rede com 22 neurônios na camada escondida foi escolhida como a melhor rede, baseando-se na sua alta taxa de acertos. No caso das redes MLP sem usar padrões obtidos com PCA ($m=75$) foi observada uma média de acertos de 53,83\% e um desvio padrão de 0,47\%; porém não foi observada uma mediana tão distante da média (mediana de 50,00\%), indicando que os resultados estavam mais simétricos. Neste caso, a rede com 19 neurônios na camada escondida foi escolhida como a melhor rede, baseando-se na sua alta taxa de acertos de 90,47\%.

Para as redes SOM-LVQ com o uso de PCA ($m=50$) foi observada uma redução na média de acertos das redes em comparação com as redes MLP. Com uma média de acertos de 50,49\% e um desvio padrão de 0,23\%. Como nas redes anteriores, foi observada uma mediana distante da média (mediana de 40,91\%), indicando que houve resultados díspares em relação à média. Contudo, o baixo valor do desvio padrão indica que os valores das taxas de acertos para as redes não oscilaram muito em relação à média. Assim, a rede com 10 neurônios na camada escondida foi escolhida como a melhor rede, com uma taxa de acerto de 66,66\%. Para as redes SOM-LVQ, sem usar padrões obtidos com PCA ($m=75$), observou-se também uma redução da efetividade de acertos da rede em comparação com as redes MLP, com uma média de acertos de 47,14\% e um desvio padrão de 0,15, com uma mediana não muito distante da média (mediana de 50,00\%) indicando que não houve resultados díspares em relação à média, embora possa ser notado com clareza que a taxa de acerto da melhor rede foi bastante alta: 81\%. Neste caso, a rede com 17 neurônios na camada escondida foi escolhida como a melhor rede, baseando-se na sua alta taxa de acertos. Depois de selecionar as melhores redes, na segunda etapa foram obtidos os resultados da Tabela \ref{tab:resultados2}.
\begin{table}[htbp]
	\centering
	\footnotesize
	\caption{Resultados da segunda etapa da classificação, usando PCA ($m=50$) e sem uso do PCA ($m=75$)}
		\begin{tabular} {c|c|c|c|c}
			{Rede} & {$m$} & {$\bar{\eta}\pm \Delta\eta$ (\%)} & {$\eta_{\mathrm{MED}}$ (\%)} & {$\eta_{\mathrm{OTM}}$ (\%)}\\
			\hline
      {\multirow{2}{*}{MLP}} & {50} & {60,95 $\pm$ 0,47} & {70,45} & {80,95}\\
      {} & {75} & {55,71 $\pm$ 0,49} & {68,18} & {80,95}\\
      \hline
      {\multirow{2}{*}{SOM LVQ}} & {50} & {52,85 $\pm$ 0,21} & {36,36} & {61,90}\\
      {} & {75} & {49,76 $\pm$ 0,15} & {45,45} & {71,42}\\      
			\hline			
		\end{tabular}
	\label{tab:resultados2}
\end{table}

Para as redes MLP usando padrões obtidos com PCA ($m=50$), foi observada uma média de acertos de 60,95\%, um desvio padrão de 0,47\%, e uma mediana de 70,45\%, indicando uma certa distância entre as taxas de acertos das redes. Isso fica mais claro quando se observa a taxa de acerto da melhor rede, de 80,95\%. No caso das redes MLP sem usar padrões obtidos com PCA ($m=75$), obteve-se uma média de acertos de 55,71\%, desvio padrão de 0,49\% e mediana de 68,18\%. A melhor rede teve uma taxa de acerto de 80,95\%. Foi obtido, assim, um resultado parcialmente similar àquele obtido para as redes com PCA, porém com uma média de acertos inferior.

Para as redes com SOM-LVQ usando padrões com PCA ($m=50$), novamente foi observada uma redução na média de acertos das redes em comparação com as redes MLP. Observou-se uma taxa de acertos de 52,85\%, um desvio padrão de 0,21\%, e uma mediana de 36,36\%. O baixo desvio padrão e o valor da mediana sugerem uma razoável assimetria entre os dados. Contudo a taxa de acerto da melhor rede não ficou tão distante da média: 61,90\%. No caso das redes SOM-LVQ sem usar padrões obtidos com PCA ($m=75$), obteve-se uma média de acertos de 49,76\%, uma média de 0,15\%, e mediana de 45,45\%. Porém, mesmo com o valor relativamente baixo de desvio padrão e a proximidade entre mediana e média, obteve-se uma melhor rede com uma taxa de acerto de 71,42\%, ficando acima da melhor rede SOM-LVQ com PCA.

\section{Discussão e Conclusões} \label{sec:quanti_conclusoes}

O uso de redes MLP no método de triagem virtual proposto neste trabalho resultou numa taxa de acerto de 81\%, o que pode ser considerado razoável, uma vez que a triagem virtual desempenha o papel de pré-classificação. Contudo, para validar a proposta, seria necessário realizar uma triagem usando especialistas humanos, especificamente patologistas. Este trabalho busca na verdade ser uma prova de conceito, mostrando que é possível aplicar a ideia de CBIR para desenvolver um sistema de triagem virtual para otimizar o trabalho de análise de imagens realizado por especialistas humanos, usando o espectro de padrões para representar as imagens, sendo este o maior diferencial deste trabalho em relação a outras aplicações de CBIR.

Os resultados obtidos indicam que o uso de PCA para reduzir a quantidade de atributos suficiente para representar os dados reduziu o custo computacional, dado que reduziu a dimensionalidade, e aumentou a taxa de acerto da classificação. Nas redes MLP houve uma melhora de aproximadamente 9\% da taxa de acerto com o uso de PCA. Já nas redes SOM-LVQ houve uma melhora de 6\% em relação à taxa de acertos com o uso de PCA.

As redes MLP se mostraram mais adequadas ao problema, com alta taxa de acerto nas melhores redes: 81\%. As redes SOM-LVQ obtiveram resultados não tão razoáveis nas suas classificações. As redes obtiveram uma melhor rede com uma taxa de 71,4\% sem usar PCA superior a melhor rede com o uso de PCA, indicando que, nesse caso, as redes SOM-LVQ necessitam de mais informações para poder ter uma melhor generalização e por conseguinte se adequar ao problema em questão.
A comparação dos resultados da triagem usando MLP e SOM mostrou que as redes MLP se adequaram melhor ao problema, com uma maior taxa de acerto.

Uma das dificuldades encontradas foi o alto custo computacional aparente necessário para obtenção dos espectros de padrões das imagens, já que era calculado o espectro para cada banda de uma imagem e isso era repetido para toda a base. Depois de gerado os vetores, o treinamento para se obter as melhores redes exigia um tempo de processamento também razoável. Como trabalho futuro podem ser investigadas arquiteturas paralelas para otimização do tempo dispendido no cálculo dos espectros de padrões, buscando paralelizar ao máximo a execução das granulometrias.

\renewcommand\refname{Referências}

\bibliographystyle{lnlm}
\bibliography{arq_bib}
\end{document}